# A Vision Language Model for Generating Procedural Plant Architecture Representations from Simulated Images

Heesup Yun[a], Isaac Kazuo Uyehara[b], Ioannis Droutsas[d], Earl Ranario[a],

Christine H. Diepenbrock[b], Brian N. Bailey[b,*], J. Mason Earles[a,c,*]

[a] Biological and Agricultural Engineering, One Shields Avenue, Davis, 95616, CA, USA

[b] Department of Plant Sciences, One Shields Avenue, Davis, 95616, CA, USA

[c] Viticulture and Enology, One Shields Avenue, Davis, 95616, CA, USA

[d] Wageningen University & Research, P.O. Box 9101, 6700 HB Wageningen, The Netherlands

*Corresponding authors

## Abstract

Three-dimensional (3D) procedural plant architecture models have emerged as an important tool for simulation-based studies of plant structure and function, extracting plant architectural parameters from field measurements, and for generating realistic plants in computer graphics. However, measuring the architectural parameters for these models at the field and population scales remains prohibitively labor-intensive. We present a novel algorithm that generates the 3D plant architecture from an image, to create a functional structural plant model from an image that reflects organ-level geometric and topological parameters, providing a more comprehensive representation of the plant's architecture. Instead of using 3D sensors or processing multi-view images with computer vision to obtain the 3D structure of plants, we proposed a method that generates token sequences containing a procedural definition of the plant architecture. This work utilized only synthetic images for training and testing, where "exact" architectural parameters were known, which allowed for testing of the hypothesis that organ-level architectural parameters could be extracted from imagery data using a vision language model (VLM). A synthetic dataset of cowpea plant images was generated using the Helios 3D plant simulator, with the detailed plant architecture encoded in XML files. We developed a plant architecture tokenizer for the XML file defining plant architecture, converting it into a token sequence that a language model can predict. Then, a VLM was trained to predict plant architecture token sequences from images. Our results demonstrated that the model can predict plant architecture tokens with an F1 score of 0.73 in a teacher-forcing scenario. Evaluation of the model was performed through autoregressive generation, achieving a BLEU-4 score of 94.00% and a ROUGE-L score of 0.5182. Our model outperformed feature regression-based methods in estimating bulk plant-level traits that require understanding of the occluded 3D structure of the plant, such as leaf count and leaf area. This led to the conclusion that such plant architecture model generation and parameter extraction were possible based on synthetic images, and thus, future work will be performed to extend the approach to real imagery data.

***Keywords***: Plant architecture, Plant modeling, Vision language model, Digital agriculture, Synthetic data

# Introduction

Understanding the three-dimensional (3D) structure of plants is necessary to understand the efficiency of plant light utilization for canopy photosynthesis and net primary production (Burgess et al., 2017; Zhang et al., 2022). Also, the 3D structure of plants plays a crucial role in quantitatively analyzing how these structural characteristics affect yield and associated relationships with genes, ultimately allowing for more efficient selection of varieties. For example, modern corn varieties have more vertical leaf angles compared to older varieties, allowing them to be planted at higher densities in a smaller area while still allowing light to reach the lower leaves (Duvick et al., 2003; Elli et al., 2023; Hammer et al., 2009). Olorunwa et al. (2022) measured plant height, node number, and leaf number, and demonstrated the effect of waterlogging on cowpea genotypes. Digrado et al. (2020) reported that canopy architectural traits contributed to 38.6% of the variance observed in the canopy photosynthesis of cowpea crops.

Plant height and stem diameter of rice plants affect the resistance to high wind, reducing damage from lodging caused by typhoons (Kashiwagi and Ishimaru, 2004; Wu et al., 2022; Zhao et al., 2023). Plant structure also affects how they compete for space with surrounding plants, as well as their growth over time (Ford, 2014).

While conventional 3D scanning of plants using LiDAR scanners or a structure from motion algorithm mainly focuses on capturing bulk plant geometry (Li et al., 2025), functional structural plant models (FSPMs) integrate a 3D representation of plant architecture at the organ level with biophysical modeling of plant function (Sievänen et al., 2014; Soualiou et al., 2021). These benefits enable the creation of a plant model that accurately reflects the characteristics of real plants and facilitate the analysis of target phenotypes of focus in breeding, which can inform the development of improved genotypes (Yu et al., 2023).

There have been various studies to represent FSPMs based on reproducible syntax. Plant modeling methods rooted in L-systems (Lindenmayer, 1968a, 1968b), such as L-Studio (Karwowski and Prusinkiewicz, 2004), L-Peach (Allen et al., 2005), and XL (Kurth and Ong, 2017), define plant architecture based on the node-edge structure of L-systems and add parameters related to plant organs. Schnepf et al. (2018) developed a method of modeling individual roots as objects using object-oriented design, and CPlantBox (Digrado et al., 2020) extended this to the entire plant, storing the plant's structure as an extensible markup language (XML) file to perform functional structural modeling and water and carbon flows simulations. Yun and Kim (2023) developed CropBox, a crop modeling software that provides a high-level approach to define FSPM using the Julia programming language. For plants with one or two branches, such as early vegetative plants, FSPMs can be tuned to closely reproduce the structure of actual plants in the field. However, defining the architecture and organ parameters of more complex and self-occluded plants becomes prohibitively difficult. Therefore, an alternative method that reduces the time and cost of generating FSPMs for a complex plant is needed.

Recent development of vision language models (VLMs) that can understand images and reproduce structured output, such as comma-separated values (CSV), XML, and programming languages, has shown feasibility for generating structured output for 3D objects and shapes. For example, Wang et al. (2024) introduced the LLaMA-Mesh algorithm, which is a vision language model that generates a sequence of natural language tokens defining 3D vertices and faces from an image. Since the methodology used in LLaMA-Mesh suggests the concept of generating language tokens defining 3D structure using conventional VLMs, it could be a potential solution that can reduce the time and cost of generating a structured plant architecture string. However, up until now, algorithms that use VLMs to generate structured output defining the plant architecture from images have not yet been tested.

Therefore, this research demonstrates the fundamental viability of generating procedurally defined plant models from plant images using a VLM, offering both the plant architecture tokenization approach and the dataset necessary for predicting and generating plant structure. The main contributions of this work are the development of a tokenization method for plant architecture models and the validation of a hypothesis that a VLM can generate plant architecture models using advanced synthetic training data that incorporates foundational knowledge of plant architectures. The use of synthetic training data not only provides an accurate benchmark for evaluating model training but also accelerates dataset generation, as obtaining such a benchmark is practically infeasible through real-world field experiments. This work lays the foundation for developing larger, modern VLMs to generate plant architecture models with more comprehensive training data in the future.

# Materials and methods

## Plant architecture model

The typical structure of a dicotyledon plant in vegetative growth consists of a shoot and a root system. This work focused on modeling the shoot system, which consists of internodes, petioles, and leaves. In reproductive growth, plants produce flowers, fruits, and seeds, changing the resource allocation and their branching pattern. In this paper, generating plants during vegetative growth was demonstrated, and generating plants in the reproductive state will be done in a future study. The basic plant architectural unit in the model is the phytomer, which consists of an internode with one or more petioles at its end, and each petiole contains one or more leaves (Teichmann and Muhr, 2015). Shoots are formed by connecting multiple phytomers end-to-end, which occurs when the vegetative bud at the shoot tip develops into a new phytomer (i.e., "growth"). Shoots can spawn lateral child shoots from vegetative buds, which are located where the internode and petiole are attached, creating a nested topological structure. For example, Fig. 1 (a) shows an actual cowpea plant, and Fig. 1 (b) represents the corresponding topological diagram. Knowing the order and relationship between each plant organ allows us to implement the actual 3D plant structure in a simulation program.

In this work, the Helios plant simulation framework was used to generate randomly varying cowpea plants based on its architectural model. Bailey (2019) developed Helios, a scalable 3D plant and environmental biophysical modeling framework. This framework is implemented in C++ and incorporates sub-models, known as plugins. Helios can simulate various types of plant biophysical processes, including energy balance, photosynthetically active radiation (PAR), and photosynthesis simulations, based on 3D geometry and ray tracing. It is possible to simulate plant structure and growth and output the full plant structure in XML file format using the plant architecture plug-in.

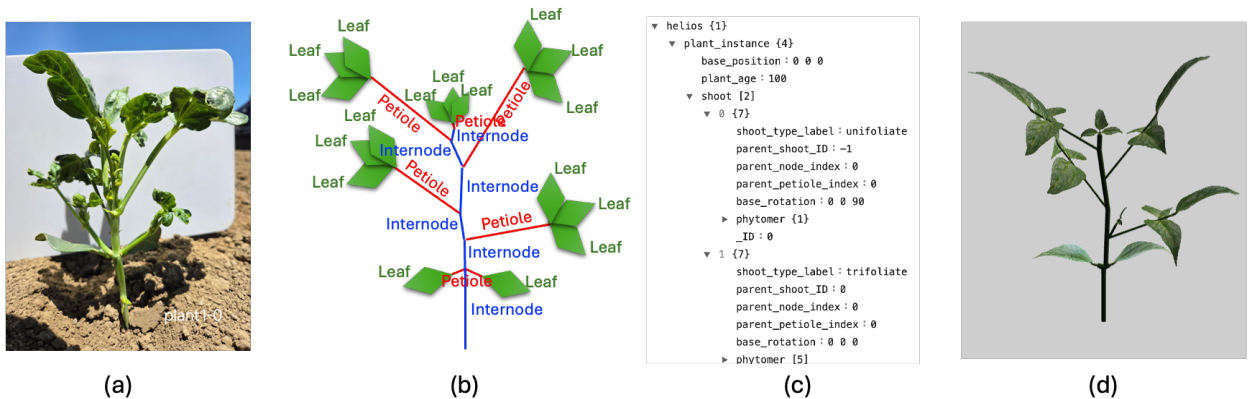

*Figure 1. Schematic depiction of how model plant architecture is defined from real plants. (a) Cowpea plant in its early growth stage (b) Simplified diagram with plant organ annotations (c) Example plant architecture XML created based on the annotations (d) Rendering of the plant architecture XML using the Helios simulator*

The plant architecture can be encoded in an XML file format, which uniquely parameterizes the topology and geometry of plant organs. The XML file contains two types of plant architectural information: the relative connectivity of plant organs (i.e., topology) and the parameters defining the geometry of each plant organ. The relative connectivity of plant organs is defined by the order in which the plant organs are defined in the XML syntax and the ID of the parent instance. For example, the parameter set defining a shoot has the shoot's unique ID and information about the parent node where the shoot is located. The parameters defining the shape of each plant organ include length, diameter, scale, angle, etc., depending on the type of plant organ, and define the individual geometric characteristics of the plant organ. Table 1 summarizes the plant architectural parameters used in the Helios plant architecture XML structure. The XML files include base positions and plant age as comprehensive plant-level parameters. The parameters defining a shoot encompass the shoot type along with its rotational parameters. The internode parameters provide geometric information, such as length, radius, pitch, and phyllotactic angle. Connecting the leaf to the internode, the petiole is characterized by parameters that specify its length, radius, pitch, curvature, and scale. Lastly, each leaf is associated with a scale parameter that determines its size, in addition to pitch, yaw, and roll angles, which indicate its angle transformation from the parent organ's coordinate system. Collectively, these elements provide a comprehensive representation of the plant's structure and are defined for each geometric element.

There are two approaches to generating the architecture of a plant via an XML file. The first approach involves manually creating the plant structure and writing it directly to an XML file, which can then be read into Helios to generate the plant geometry. Fig. 1 (c) is an example of a plant architecture XML file based on a real plant, and Fig. 1 (d) shows the rendering of this using the Helios program. The second way is to generate the plant architecture dynamically and procedurally based on the Helios plant architecture plug-in. The plug-in can automatically generate a plant architecture model of a specific plant growth stage, simulate plant growth, and modify the plant architecture model by pre-defined rules and random sampling from the parameter distribution of plant organs. Therefore, even with the same plant age and base rotation, generated plant architectures have different branching structures and shapes.

## Synthetic cowpea dataset

It is necessary to have ground-truth annotations of the plant architecture corresponding to the plant images to predict the plant architecture from images using VLMs. However, acquiring a large-scale agricultural image dataset with high-quality annotations is costly and time-consuming, which can limit algorithm development (Dyrstad and Øye, 2025). To overcome these difficulties, synthetic data generated by simulation programs can be utilized.

A dataset consisting of 3D cowpea plant models was procedurally generated based on randomly varying architectural parameters following a specified distribution (Table 1). The dataset was generated using the procedural plant architecture model in Helios v1.3.14, which is described in more detail in the previous section. During the dataset generation process, consistent random seeds were assigned to each plant, based on the generated plant index for reproducibility (Fig. 2 (a)). The growth of the crop was simulated by creating a plant on day 0 using the Helios Cowpea library and then advancing it over time to day 39 (Fig. 2 (b) and (c)). All simulated cowpea plants in this dataset were in the vegetative growth stage, simplifying the problem definition by generating a plant architecture model without reproductive organs. The distributions of plant organ parameters are defined within the Helios library and can be customized by the user. For the generation of this dataset, the default parameter distributions defined in the library were used. As a result, the generated plant architecture dataset represents an imaginary population of cowpea plants that were sampled from specific parameter distributions. 10,000 random seeds were used to generate the day 0 plant, and plant growth was simulated on a daily basis up to day 39. Therefore, a total of 10,000x40 = 400,000 simulated images and plant architecture XML pairs were generated. Additionally, for multi-view dataset generation, images at azimuthal viewing angles of 0°, 120°, and 240° were also rendered for each XML file (Fig. 2 (d)).

*Table 1 List of plant architectural parameters from Helios plant architecture plug-in. The default values and parameter distributions are used from Helios Cowpea library. More detailed information can be found here;[https://baileylab.ucdavis.edu/software/helios/_plant_architecture_doc.html](https://baileylab.ucdavis.edu/software/helios/_plant_architecture_doc.html)*

| Parameter | Description | Values and distributions | Units |
|---|---|---|---|
| base_rotation_pitch | Plant rotation (pitch) | Uniform(0, 10) | deg. |
| base_rotation_yaw | Plant rotation (yaw) | Uniform(0, 360) | deg. |
| base_rotation_roll | Plant rotation (roll) | Uniform(0, 360) | deg. |
| plant_age | Days after germination | Uniform(0, 39) | days |
| shoot_type_label | Shoot types based on leaf composition | Unifoliate: Constant(1) Trifoliate: Constant(3) | index |

| Parameter | Description | Values and distributions | Units |
|---|---|---|---|
| shoot_base_pitch | Shoot base pitch angle | Unifoliate: Constant(40)<br>Trifoliate: Uniform(40, 60) | deg. |
| shoot_base_yaw | Shoot base yaw angle | Unifoliate: Uniform(0, 360)<br>Trifoliate: Uniform(-20, 20) | deg. |
| shoot_base_roll | Shoot base roll angle | Constant(90) | deg. |
| internode_length | Initial internode length | Constant(0.002) | m |
| internode_radius | Initial internode radius | Constant(0.0015) | m |
| internode_pitch | Internode pitch angle | Unifoliate: Constant(0)<br>Trifoliate: Constant(20) | deg. |
| internode_phyllotactic_angle | Phyllotactic angle | Uniform(145, 215) | deg. |
| petiole_length | Petiole length | Unifoliate: Constant(0.0004)<br>Trifoliate: Uniform(0.06, 0.08) | m |
| petiole_radius | Petiole radius | Unifoliate: Constant(0.0001)<br>Trifoliate: Constant(0.0018) | m |
| petiole_pitch | Petiole pitch angle | Unifoliate: Uniform(60, 80)<br>Trifoliate: Uniform(45, 60) | deg. |
| petiole_curvature | Petiole curvature | Trifoliate: Uniform(-200,-50) | deg. |
| leaflet_scale | Leaflet size scale | Trifoliate: Constant(0.9) | |
| leaf_scale | Leaf size scale | Unifoliate: Constant(0.02)<br>Trifoliate: Uniform(0.09,0.12) | |
| leaf_pitch | Leaf pitch angle | Unifoliate: Uniform(-10, 10)<br>Trifoliate: Normal(45, 20) | deg. |
| leaf_yaw | Leaf yaw angle | Unifoliate: Constant(0)<br>Trifoliate: Constant(10) | deg. |
| leaf_roll | Leaf roll angle | Constant(-15) | deg. |

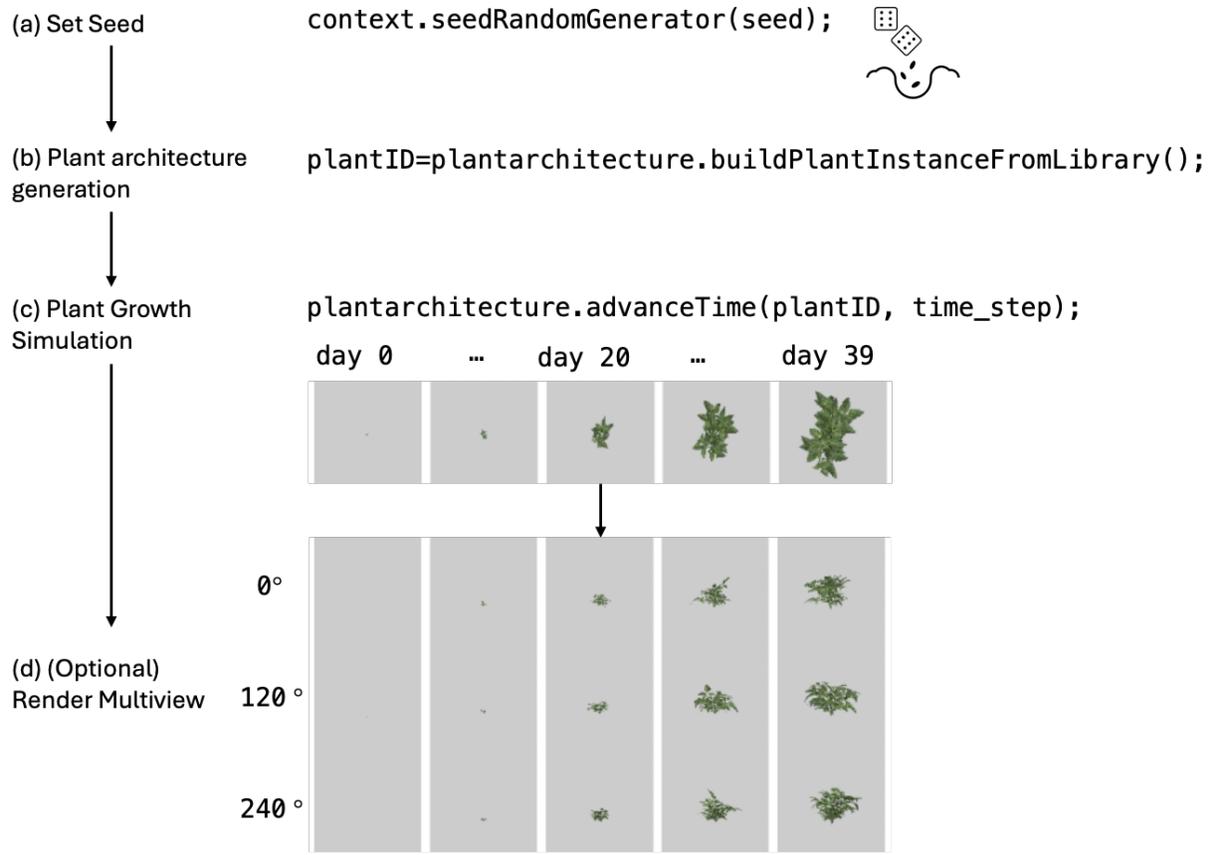

*Figure 2. Process of generating a synthetic cowpea plant architecture dataset. We set the random seed (a) for repeatability and generated plant architecture (b). (c) Then, we advanced time to simulate plant growth (d), and optionally rendered side-view images.*

## Plant architecture tokenizations

Helios can encode and decode XML files (Fig. 1(c)), a structured document format that enables the export or import of plant architecture models to interface with the Helios program. While this text-based format can be used to train a VLM, processing XML files requires an intermediate step that converts the structured text into embedding vectors from which the model can learn. However, applying a conventional text tokenization method to an XML text is inefficient because the XML file structure and elements, such as opening and closing tags with special characters, unnecessarily increase the sequence length that requires a larger context length for language models, resulting in a larger memory footprint and reduced generation performance.

Therefore, a new method was needed to streamline the XML file to essential plant architecture model information. In this paper, a specialized plant architecture tokenizer was developed, representing the plant organs and parameters as tokens.

The XML file contains information on a series of plant organs with a hierarchical structure, which can be represented by their organ token and the following parameter tokens using the plant architecture tokenizer. The plant organ tokens include the branching order in which the organ is attached, as well as the type of organ. This preserves the hierarchy and sequential relationship between plant organs and makes the plant architecture token sequence semantically identical with the original XML file, which enables conversion between the two formats without information loss.

Fig. 3 illustrates an example of tokenizing plant architecture into a sequence of tokens. The first shoot tag is tokenized as "00", where the leading 0 indicates the primary branch, and the trailing 0 represents the shoot organ. Next, the phytomer structures of internode, petiole, and leaves appear repeatedly. Similarly, the trailing number represents 1: internode, 2: petiole, 3: leaf. Therefore, the internode connected to the primary branch is tokenized as "01," and the petiole attached to this internode is tokenized as "02". Since the initial leaves connected to the primary branch are unifoliate, each leaf is attached to a single petiole, which is tokenized to "03". The shoot organ corresponding to the secondary branch connected to internode "01" is tokenized as "10", and the internode and petiole connected to this node are tokenized as "11" and "12", respectively. Since the leaves attached to the petioles after the first unifoliate leaves have a trifoliate configuration, each leaf is tokenized as "13", "14", and "15". Parameter tokens defining the plant organ's geometry follow the plant organ token, represented as p0 through p3 or p4, depending on the number of parameters per plant organ. This rule is applied recursively until all the plant organs are described. Combining all of this together, the plant architecture defined in Fig. 3(a) can be tokenized as shown in Fig. 3(d).

After converting the plant architecture XML to a token sequence, it is converted to categorical variables, which are token IDs that the transformer can predict, using a table shown in Fig. 3 (c). First, for the plant organ token, the branching order was multiplied by 6, and the organ type was added to make the token IDs. Next, the parameter values are converted to the token IDs. Because conventional tokenizers don't have specific token dictionaries for decimal point values, they can't convert the plant parameter token value into a single token ID. Therefore, a quantization method was applied to convert the parameter values to discrete integer values between 24 and 222, considering varying scales and distributions. To ensure a similar resolution across various scales and reduce the size of token dictionaries, combinations of a set of constant values and linearly spaced float grids were applied. For example, 0, -10, 10, and 90 degrees are frequently observed numbers in angle parameters. Also, 1 and 3 are used to represent the shoot organ that has unifoliate leaves or trifoliate leaves. As a result, those constant values are added to the mapping between parameter token values to token IDs. Besides constant values, linearly spaced grids were applied. This includes angle representations of [-40, 360] in degrees in 2.5-degree resolution. The set of decimal values was created by defining 10 evenly spaced samples within the ranges [0.1, 1.0], [0.01, 0.1], [0.001, 0.01], and [0.0001, 0.001] in meters to represent organ parameters with small decimal point values. This set of constant values and float value grids was concatenated and used to find the closest index of a certain parameter value, which can be mapped to a token ID.

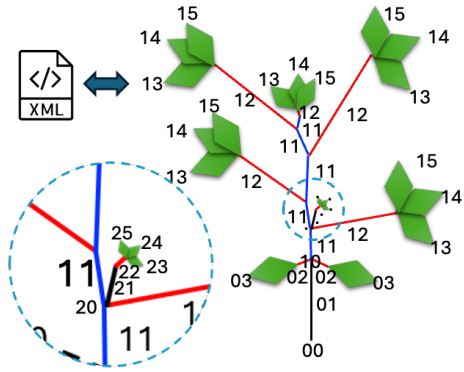

(a) Example plant organ labels  (b) Plant token annotation rules  (c) *Token IDs*

*Format: MN p0 p1 p2 p3 p4*

Branching order (M)
0: Primary branch
1: Secondary branch
2: Tertiary branch

Organ (N)
0: shoot
1: internode
2: petiole
3, 4, 5: leaf

Parameters
p0 ... p4: organ parameters

| Branching order (M) | Organ (N) | IDs (6*M + N) |
|---|---|---|
| 0 - 3 | 0 - 5 | 0 - 23 |
| Parameters | | IDs |
| Quantized floats values | | 24 - 222 |
| Special Tokens | | IDs |
| SOS | | 223 |
| META | | 224 |
| PAD | | 225 |
| EOS | | 226 |

(d) *Tokenized plant architecture (1x201)*  * **Bold**: Organ tokens and special tokens / Plain: Parameter tokens

**00**,p0,p1,p2,p3,p4,**01**,p0,p1,p2,p3,**02**,p0,p1,p2,p3,p4,**03**,p0,p1,p2,p3,**10**,p0,p1,p2,p3,p4,**11**,p0,p1,p2,p3,**12**,p0,p1,p2,p3,p4,**13**,p0,p1,p2,p3,**14**,p0,p1,p2,p3,**15**,p0,p1,p2,p3,**20**,p0,p1,p2,p3,p4,**21**,p0,p1,p2,p3,**22**,p0,p1,p2,p3,p4,**23**,p0,p1,p2,p3,**24**,p0,p1,p2,p3,**25**,p0,p1,p2,p3,**11**,p0,p1,p2,p3,**12**,p0,p1,p2,p3,p4,**13**,p0,p1,p2,p3,**14**,p0,p1,p2,p3,**15**,p0,p1,p2,p3,**11**,p0,p1,p2,p3,**12**,p0,p1,p2,p3,p4,**13**,p0,p1,p2,p3,**14**,p0,p1,p2,p3,**15**,p0,p1,p2,p3,**11**,p0,p1,p2,p3,**12**,p0,p1,p2,p3,p4,**13**,p0,p1,p2,p3,**14**,p0,p1,p2,p3,**15**,p0,p1,p2,p3,**11**,p0,p1,p2,p3,**12**,p0,p1,p2,p3,p4,**13**,p0,p1,p2,p3,**14**,p0,p1,p2,p3,**15**,p0,p1,p2,p3,**02**,p0,p1,p2,p3,p4,**03**,p0,p1,p2,p3

(e) Plant metadata and *special tokens added (1x208)*

**SOS**,**META**,p0,p1,p2,**META**,**00**,p0,p1,p2,p3,p4,**01**,p0,p1,p2,p3,**02**,p0,p1,p2,p3,p4,**03**,p0,p1,p2,p3,**10**,p0,p1,p2,p3,p4,**11**,p0,p1,p2,p3,**12**,p0,p1,p2,p3,p4,**13**,p0,p1,p2,p3,**14**,p0,p1,p2,p3,**15**,p0,p1,p2,p3,**20**,p0,p1,p2,p3,p4,**21**,p0,p1,p2,p3,**22**,p0,p1,p2,p3,p4,**23**,p0,p1,p2,p3,**24**,p0,p1,p2,p3,**25**,p0,p1,p2,p3,**11**,p0,p1,p2,p3,**12**,p0,p1,p2,p3,p4,**13**,p0,p1,p2,p3,**14**,p0,p1,p2,p3,**15**,p0,p1,p2,p3,**11**,p0,p1,p2,p3,**12**,p0,p1,p2,p3,p4,**13**,p0,p1,p2,p3,**14**,p0,p1,p2,p3,**15**,p0,p1,p2,p3,**11**,p0,p1,p2,p3,**12**,p0,p1,p2,p3,p4,**13**,p0,p1,p2,p3,**14**,p0,p1,p2,p3,**15**,p0,p1,p2,p3,**02**,p0,p1,p2,p3,p4,**03**,p0,p1,p2,p3,**EOS**

(f) *Token IDs (1x208)*

**223**,**224**,72,60,62,**224**,**0**,40,40,114,82,77,**1**,51,41,40,139,**2**,43,41,107,133,77,**3**,52,80,40,34,**6**,40,40,40,82,79,**7**,50,41,80,40,**8**,52,40,106,118,76,**9**,51,39,82,34,**10**,52,81,40,34,**11**,51,79,36,34,**12**,90,40,40,82,79,**13**,45,40,40,40,**14**,43,40,40,118,76,**15**,45,39,82,34,**16**,45,81,40,34,**17**,45,79,36,34,**7**,50,41,40,150,**8**,52,40,96,118,76,**9**,51,39,82,34,**10**,52,81,40,34,**11**,51,79,36,34,**7**,50,41,36,150,**8**,52,40,100,118,76,**9**,51,39,82,34,**10**,52,81,40,34,**11**,51,79,36,34,**7**,50,41,40,150,**8**,52,40,100,118,76,**9**,51,39,82,34,**10**,52,81,40,34,**11**,51,79,36,34,**7**,45,41,82,24,**8**,41,40,100,118,76,**9**,50,39,82,34,**10**,50,81,40,34,**11**,50,79,36,34,**2**,43,41,107,133,77,**3**,52,38,40,34,**226**

*Figure 3.* An example of plant architecture annotations and tokenizations. (a) shows a simplified plant architecture diagram based on an input XML file and (b) plant organ annotation rules. The Table (c) shows the rules converting plant organ annotations and float values of plant organ parameters to unique integer values. (d) shows plant architecture tokens of example plant architecture, which can also be represented in XML file format. (e) shows the sequence with special tokens for VLM training. (f) shows the token IDs that the model actually predicts as categorical variables.

Since the plant size information was lost if the input image was cropped and resized during the image preprocessing step, the plant's width and height in meters in the original image were estimated and concatenated in the input token string. Additionally, the vegetation fraction, which represents the proportion of the image window covered by vegetation, was also calculated and included as plant metadata. This plant metadata was encapsulated with <META> tokens and added at the start of the plant architecture tokens. Next, a <SOS> token indicating the beginning of the sentence and an <EOS> token indicating the end of the sentence were added to the beginning and end of the sequence. Fig. 3(e) shows an example with plant metadata tokens, including start and end tokens, and Fig. 3(f) shows the final input token IDs, ready for training the language model decoder.

## Algorithm structure

After tokenizing the plant architecture XML into a token sequence, the model was trained to perform the next token prediction (NTP) task in a teacher-forcing scenario. Fig. 4 shows the overall algorithm structure. As mentioned in the previous section, the input image was cropped around the plant and resized to $224^2$ or $448^2$ based on the model configuration. In addition, the plant size information is quantized and concatenated to the beginning of the input token IDs. The NTP tasks estimated the next token starting from the SOS token and plant metadata tokens to complete the plant architecture token sequence. The target token sequence was one token shifted to predict the start of the metadata token to the EOS token.

The model and training code were implemented in Python programming language based on the Hugging Face (https://huggingface.co) library's vision encoder-decoder model, which consists of layers of transformer encoder and decoder blocks (Vaswani et al., 2017; Dosovitskiy et al., 2021). A vision foundation model DINOv2 (Oquab et al., 2024) was utilized with its weights frozen to serve as a feature extractor, thereby maintaining generalization performance and reducing training time. The vision encoder's patch embedding sequence was connected to the decoder's cross-attention layer to perform the NTP task with a vision context. The decoder block's configuration follows the GPT-2 model series.

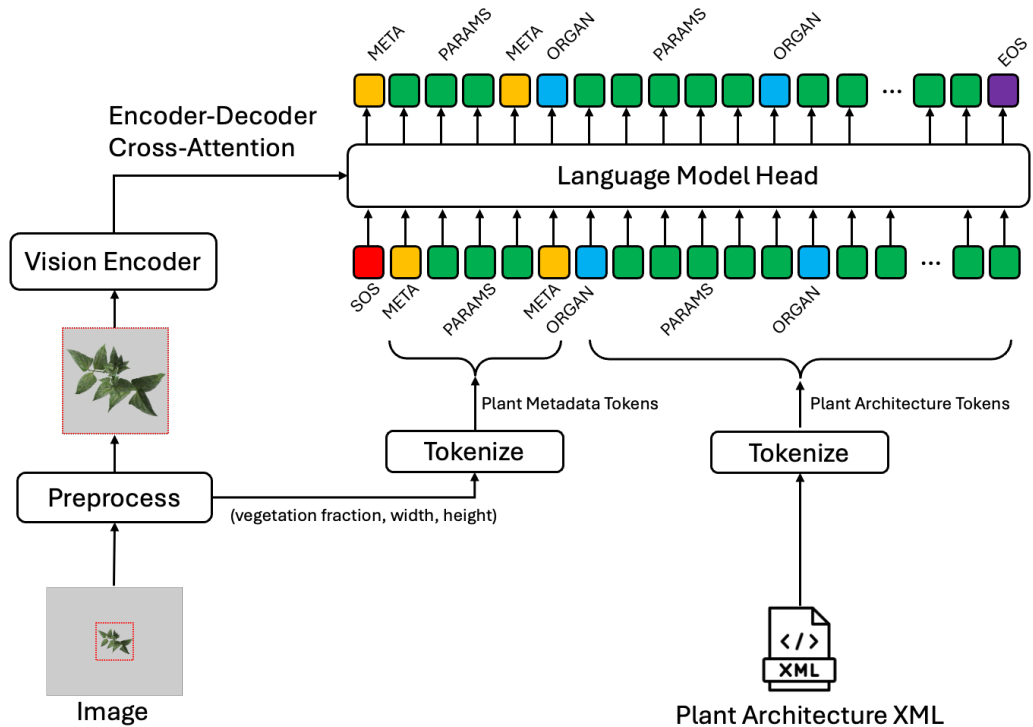

*Figure 4. Proposed algorithm structure for the next token prediction task from an image. A plant architecture XML was visualized by a plant simulator to create an input image for the model, and tokenized to provide a ground truth token sequence. The preprocess block cropped the input image around the plant, resized it to the* vision encoder input size, and saved the original plant size information, which is plant metadata. The *tokenize block for plant metadata converts* plant metadata *to input token IDs and concatenates them after the SOS token (red). The vision encoder block extracted visual features from the preprocessed image and was connected to the encoder-decoder cross-attention layer of the language model decoder. The target sequence was one token shifted, and the EOS token (purple) was added at the end of the sequence to train the language model decoder in a teacher-forcing scenario.*

## Model training procedure

To examine the performance differences based on the number of parameters in the vision encoder and language decoder within the vision encoder-decoder model, two combinations of vision encoders (ViT-S, 21M parameters; and ViT-B, 86M parameters) and two language decoders (gpt-2, 115M parameters; and gpt-2-medium, 406M parameters) were tested. Additionally, to understand how the amount of information affects plant architecture generation performance, training was conducted using both top-view images and multi-view images, which include side-view images, with combined input images resized to 224 and 448 pixels in width and height for the training dataset. Consequently, a total of 16 different model combinations were trained, and their evaluation metrics were calculated.

During the training process, 80% of the dataset was used as training data, 10% as validation data, and 10% was reserved as a test dataset. To increase stability in the early stages of training (Hägele et al., 2024), the learning rate was gradually increased from 0 to $10^{-4}$, reaching 20% of the total training steps, and then linearly decreased to 0 for the remaining steps to complete the training. The training batch size was 16, and the gradient accumulation size was set to 4, resulting in an effective batch size of 64. Training was conducted to minimize the cross-entropy loss function and was carried out on an NVIDIA A100 GPU for 4 epochs.

The accuracy and weighted F1 score were monitored in a teacher-forcing scenario during the model training, where the model estimates the $n^{th}$ token given the previous true $(n-1)^{th}$ tokens. The training accuracy of plant architecture tokens was calculated using the numbers of true positives (TP), true negatives (TN), false negatives (FN), and false positives (FP) between the predicted $n^{th}$ token from 1 to $(n-1)^{th}$ token sequence and the true $n^{th}$ token.

$$Accuracy = \frac{TP + TN}{TP + TN + FN + FP}$$

The weighted F1 score was calculated as the harmonic mean of precision and recall weighted by the percentage of token IDs. For example, for a token k, the precision and recall can be calculated below

$$Precision_k = \frac{TP_k}{TP_k+FP_k},$$

$$Recall_k = \frac{TP_k}{TP_k+FN_k},$$

$$F1_k = \frac{2 \times Precision_k \times Recall_k}{Precision_k+Recall_k}.$$

The weighted F1 score was calculated using the weighted F1 score equation

$$Weighted\ F1 = \frac{1}{\sum_{k=0}^{227} n_k} \sum_{k=0}^{227} n_k F1_k.$$

## Model evaluation metrics

In the model evaluation stage, the trained model generates plant architecture tokens from given SOS and plant metadata tokens with image context, also known as auto-regressive generation. Fig. 5 shows the overall evaluation process. As shown in Fig. 5 (b), the predicted token was then appended to the input token sequence, and this process continued until the EOS token was generated, thereby completing the plant architecture token sequence. A beam search with a beam size of 6 was employed to improve the generation result.

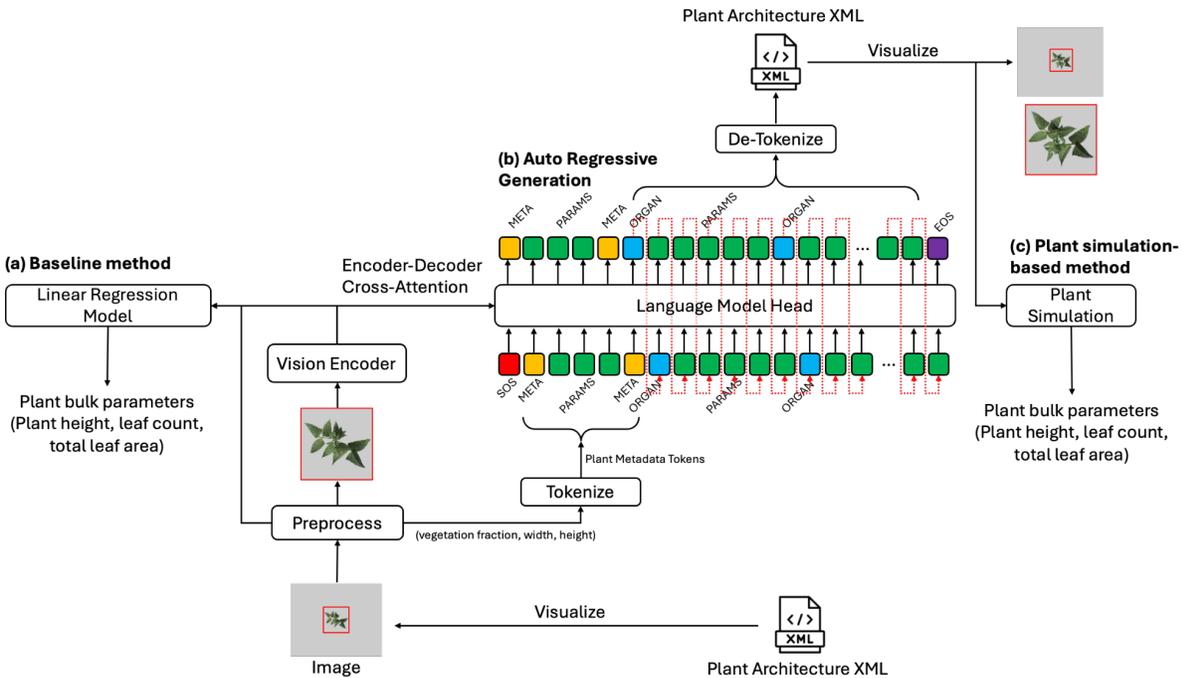

*Figure 5.* Visualization of the model evaluation process. Red dotted arrows in (b) auto regressive generation mean the generated token will be appended to the end of the input token sequence. Generated plant architecture XML is visualized for qualitative evaluation. Plant bulk parameters are estimated in (a) the baseline method based on linear regression, and (c) the plant simulation-based method. Arrows flow in (a) baseline method that combines image features and plant metadata values. The arrow connected to the plant simulation represents a de-tokenized and recovered plant architecture XML file, which can be used for plant architecture model simulation.

Unlike the teacher-forcing training, the generation results have varying lengths from the ground truth, resulting in different numbers of plant organs. Therefore, it is necessary to use evaluation metrics for comparing two sequences of different lengths. To do this, the BLEU-4 score (Papineni et al., 2001) and ROUGE-L (Lin, 2004) were calculated.

The bilingual evaluation understudy (BLEU) score is a precision-based metric comparing token sequences between the generated and ground truth. Even if the metric is developed for machine translation evaluation, it can be extended to compare any two token sequences. The BLEU score counts how many n-gram patterns in a generated token sequence also appear in the ground truth token sequence. The N-grams are groups of N consecutive tokens within a token sequence. The 1-gram BLUE score calculates the ratio of tokens in the generated token sequence that are also present in the ground truth. Since plant organ tokens (i.e., "00", "01", "02", … "35") appeared at least once in the generated token sequence, the ratio of plant organ token in the generated token sequence, approximately 20% (1/5) between 17% (1/6) will be naive baseline for 1-gram matching score if the generated token sequence have all the plant organ tokens but wrong plant parameter tokens. Similarly, if a correct combination of plant organ tokens with three organ parameter tokens, or any consecutive combinations of four tokens, also appear in ground truth, it is counted as a 4-gram matching.

The BLEU-4 score is calculated by averaging 1-gram, 2-gram, 3-gram, and 4-gram matching scores. A higher BLEU-4 score implies that the generated tokens have common patterns with the ground truth. In other words, a lower BLEU-4 score means that the generated token sequence has patterns that do not exist in the ground truth. For example, if the BLEU-4 score is 100%, it indicates a sufficient condition that all the patterns of the generated plant architecture tokens are present in the ground truth plant architecture token sequence. It also means that if the model generated only one branch with correct patterns, it can still show a high precision score. To prevent this, a brevity penalty is applied to penalize the sequence length of the generation result if it's shorter than the ground truth.

$$BLEU4 = \left(1, \exp\exp\left(1 - \frac{\sum l_{GT}}{\sum l_{gen}}\right)\right)\left(\prod_{n=1}^{4} n-gram\ BLUE\right)^{1/4}$$

ROUGE-L calculates the longest common subsequence (LCS) between the generated token sequence and the ground truth token sequence, and uses this to compute the F score. The $F_{LCS}$ score for the i[th] sentence is calculated as follows:

$$R_{LCS} = \frac{LCS(Gen,GT)}{len(Gen)},$$

$$P_{LCS} = \frac{LCS(Gen,GT)}{len(GT)},$$

$$F_{LCS} = \frac{(1+\beta^2)R_{LCS}P_{LCS}}{R_{LCS} + \beta P_{LCS}}, where\ \beta = P_{LCS}/R_{LCS}.$$

The ROUGE-L score is the average of the $F_{LCS}$ scores for N pairs of generated plant architecture token sequences and their corresponding ground truth plant architecture token sequences. A higher ROUGE-L score indicates that longer common token patterns are found between the generated plant architecture token sequence and the ground truth plant architecture token sequence. If the ROUGE-L score is 1.0, it means that the generated plant architecture token sequence is the same as the ground truth plant architecture token sequence. Even if the ROUGE-L score can show low scores for inverted sequences (i.e., same plant architecture with a swapped secondary branch) and paraphrased sequences (i.e., visually similar plant architecture but has negligible token offsets for every organ parameter), the score can be a metric that shows structural similarity and preservation of pivotal plant architecture patterns.

The evaluation used 4,000 images from the dataset, which were not used during the training or validation step. The evaluation step was done after finishing the model training and validation for each model configuration.

## Plant architecture parameter evaluations

After the model evaluation, the best-performing model that achieved the highest BLEU-4 and ROUGE-L scores, as indicated in Table 2, was selected for a detailed evaluation in terms of plant architecture. First, we compared the tokens related to the plant's base rotation angles and plant age from the generated and the ground truth plant architectures. The base rotation angles are the primary branch's rotation angles, which determine the plant's overall rotation and other plant organs attached to it. The plant age indicates the number of days since germination in the simulation. In addition to the direct parameter comparison above, the distributions of the other organ parameters were compared to determine whether the generated plant architecture organ parameters followed the distribution of the ground truth. The parameter distributions were plotted as histograms, and the Wasserstein distance (WD) was computed to measure the differences between the two distributions. Wasserstein distance is a method for measuring the distance between two probability distributions, defined as follows

$$W_p(P_r, P_g) = \left(E_{(x,y)\sim\gamma}[d(x,y)^p]\right)^{1/p}.$$

When p = 1, it is referred to as the earth mover's distance, and a smaller value indicates that the two distributions can be matched at a lower cost.

## Plant simulation-based trait evaluation

Plant height, leaf count, leaf area, and leaf angle distribution are important factors that influence the photosynthetic and water-use efficiency of plants, also affecting yield and environmental resistance. For instance, Truong et al. (2015) utilized FSPM to demonstrate that differences in leaf angles can affect the amount of photosynthesis, which in turn can influence the yield of sorghum. Pele et al. (2016) reported that plant height, number of leaves per plant, stem diameter, and root dry mass of cowpea affect the drought susceptibility index.

Calculating the number of plant organs, such as leaf count and number of nodes, can be done by analyzing the plant architecture sequence. However, estimating plant architectural traits such as plant height and leaf angle distribution directly from plant architecture sequence is not possible because tokens are the basic building blocks of plant structure, not a simulated plant model. The angle of each leaf is influenced by the series of coordinate transforms defined by parameter tokens of the internode and petiole from the first shoot to the leaf. Similarly, plant height is determined by a series of coordinate transforms and measured by the highest point above the ground. Therefore, the plant architecture sequences were loaded into a 3D plant model using a simulation program, and the final values were computed afterward (Fig. 5 (c)).

A linear regression-based baseline method was tested to evaluate the accuracy of plant height, leaf count, and total leaf area that were indirectly calculated through plant architecture model simulation. The baseline method for predicting plant traits from an image was fitting linear regression models from image features and plant metadata (Fig. 5 (a)). While the regression models cannot predict the entire complex plant architecture token sequence, traits that represent a plant's overall visual traits, such as plant height, leaf count, and leaf area, can be predicted with relatively high accuracy (Schiller et al., 2021; Singh et al., 2023; Zheng et al., 2022). Therefore, linear regression models based on fully connected layers, using the ViT-B model as a feature extractor, were fitted to predict plant height, leaf count, and total leaf area as a baseline.

# Results

## Model training and evaluation results

A total of 16 combinations of settings were trained, including view information (top-view vs. multi-view), input image size ($224^2$ vs. $448^2$ pixels), size of vision encoder model (ViT-S vs. ViT-B), and size of text decoder model (gpt-2 vs. gpt-2-medium). Table 2 summarizes the training metrics, showing the accuracy and weighted F1 score. Overall, both the weighted F1 scores showed values ranging from 0.72 to 0.73. In particular, the weighted F1 score and accuracy showed similar results regardless of the model size and input images. This indicates that, given the current data and model architecture under teacher forcing conditions, the maximum token prediction accuracy was approximately 0.73.

*Table 2 Training metrics for models with different encoder and decoder setups, across various viewing angles and input image sizes. The highest F1 score and accuracy are shown in **bold**.*

|   |   |   | Text Decoder | |
|---|---|---|---|---|
|   |   |   | gpt-2 | gpt-2-medium |

| View | Input Size | Vision Encoder | Accuracy | F1 | Accuracy | F1 |
|---|---|---|---|---|---|---|
| Top-view only | $224^2$ | ViT-S | 0.7277 | 0.7285 | 0.7282 | 0.7290 |
| | | ViT-B | 0.7260 | 0.7271 | 0.7291 | 0.7309 |
| | $448^2$ | ViT-S | 0.7269 | 0.7271 | **0.7301** | 0.7317 |
| | | ViT-B | 0.7276 | 0.7286 | 0.7284 | 0.7294 |
| Multi-view | $224^2$ | ViT-S | 0.7270 | 0.7278 | 0.7287 | 0.7288 |
| | | ViT-B | 0.7297 | **0.7329** | 0.7288 | 0.7290 |
| | $448^2$ | ViT-S | 0.7286 | 0.7292 | 0.7297 | 0.7306 |
| | | ViT-B | 0.7283 | 0.7296 | 0.7290 | 0.7303 |

As shown in Table 3, the combination of larger models in the auto-regressive generation process demonstrated better performance metrics. The BLEU-4 score was highest at 94.00% when utilizing the most model parameters (ViT-B with gpt-2-medium) with the most information (448 input size, multi-view). In other words, 94.00% of the generated plant architecture token patterns were found in the ground truth plant architecture token corpus, whereas the remaining 6.0% existed in generated sequences only. The size of the model and the amount of information did not always correlate with BLEU-4 scores. For example, when a 224-sized multi-view image was input into the ViT-B and gpt-2 combination, the score was lower than in other less favorable situations.

Similarly, in the ROUGE-L score, the highest score of 0.5182 was observed in the model with the highest BLEU-4 score. However, it was also evident that the ROUGE-L score consistently increased when larger image sizes were used, when multi-view data was provided, when a larger vision encoder was used, and when a larger text decoder was employed. However, there was no substantial difference in performance across different models. If further training is conducted, the score gap would increase; however, the evaluation results suggested that the prediction performance of various models with the current dataset did not exhibit a noticeable difference in generation quality based on these differences.

*Table 3 Model evaluation metrics for different configurations. The highest BLEU-4 and ROUGE-L scores are shown in **bold**.*

| View | Input Size | Vision Encoder | Text Decoder | | | |
|---|---|---|---|---|---|---|
| | | | gpt-2 | | gpt-2-medium | |
| | | | BLEU-4 (%) | ROUGE-L (0 - 1) | BLEU-4 (%) | ROUGE-L (0 - 1) |
| Top-view only | $224^2$ | ViT-S | 92.35 | 0.4998 | 90.67 | 0.5056 |
| | | ViT-B | 87.07 | 0.5023 | 87.85 | 0.5062 |
| | $448^2$ | ViT-S | 84.97 | 0.4996 | 85.34 | 0.5070 |
| | | ViT-B | 90.73 | 0.5030 | 90.79 | 0.5082 |
| Multi-view | $224^2$ | ViT-S | 89.52 | 0.5093 | 91.10 | 0.5154 |
| | | ViT-B | 84.88 | 0.5112 | 87.96 | 0.5150 |
| | $448^2$ | ViT-S | 85.39 | 0.5090 | 84.27 | 0.5175 |
| | | ViT-B | 89.08 | 0.5126 | **94.00** | **0.5182** |

Fig. 6 shows rendered images from a generated plant architecture sequence from the best-performing model (multi-view, 448, ViT-B, and gpt-2-medium), after converting the generated plant architecture token sequence from the input image back into XML and rendering it.

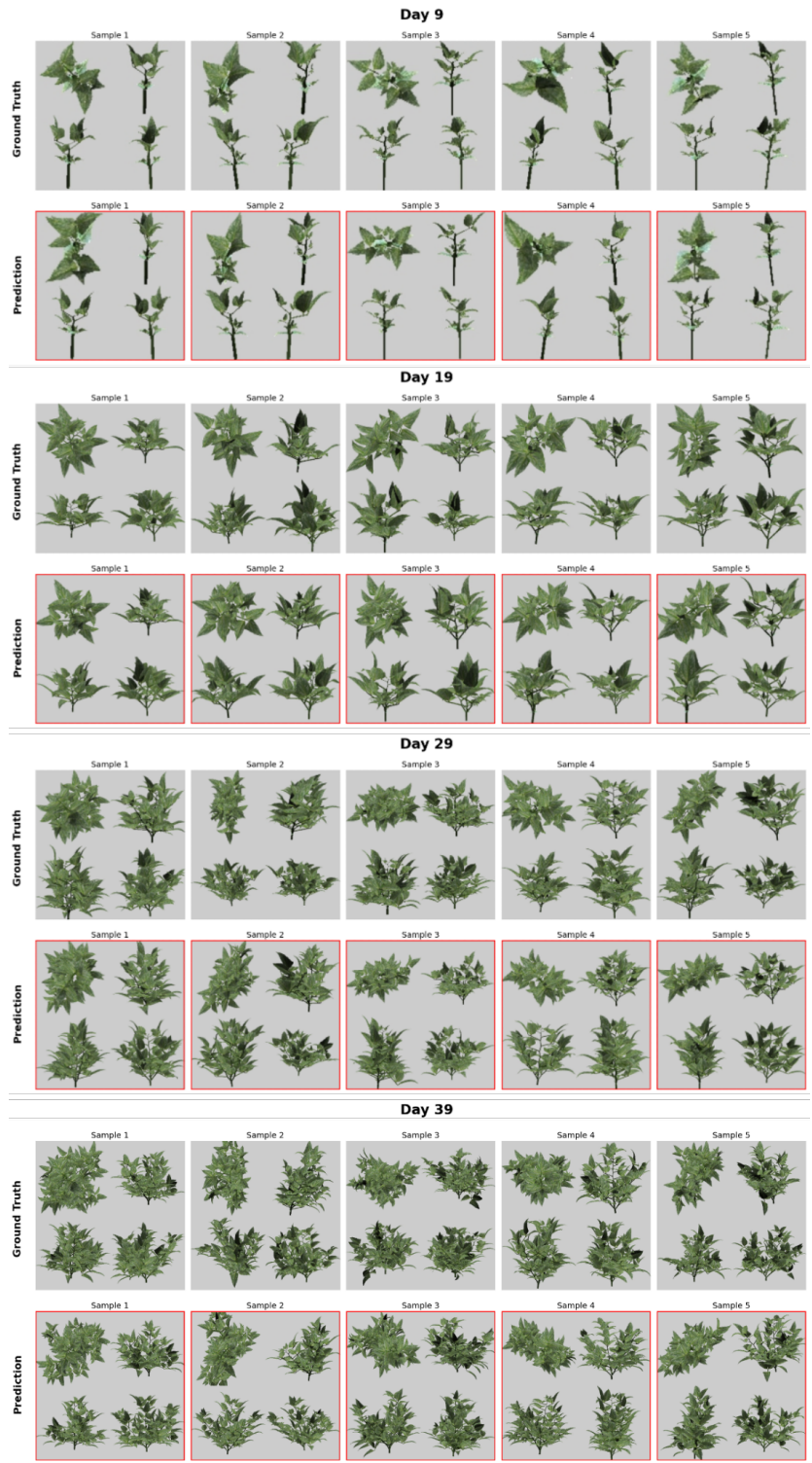

*Figure 6. Examples of rendered images of plant architecture tokens generated from the best-performing model showing the highest BLEU-4 and ROUGE-L scores.*

## Generated plant architecture token sequence evaluations

To examine how precisely the model generates the plant architecture parameters, the pitch, yaw, and roll angles of the primary branch were compared between the generated plant architecture token sequence and the ground-truth. Due to the conversion from parameter to token IDs, which converts float values into discrete integers, the parameter values exist in clusters. For example, in Fig. 7 (a), the base rotation pitch sampled from a [0, 10] uniform distribution is quantized to discrete float values ranging from 0 to 1.0 and 2.5-degree angle steps, resulting in a grid-like scatter plot. The discrete patterns introduced by tokenization in yaw and roll angle of the primary branch, which range from 0 to 360 degrees, were less pronounced (Fig. 7 (b) and (c)). The range of pitch angle of the primary branch was narrow ([0, 10]) in the synthetic dataset and showed a subtle effect on the image, making its prediction challenging. The yaw and roll angles of the primary branch were closely clustered to the y=x line; however, there are clusters observed in y=x-180 and y=x+180, suggesting the model has difficulty with 180-degree rotations around the element's roll axis. Fig. 7 (d) compares the plant age from the generated plant architecture token with the plant age from the ground-truth. The plant age prediction showed an $R^2$ value of 1.00, with a root mean squared error (RMSE) of 0.79 days. From this, it can be assumed that predicting the plant age is a relatively easier task than predicting the plant's rotation.

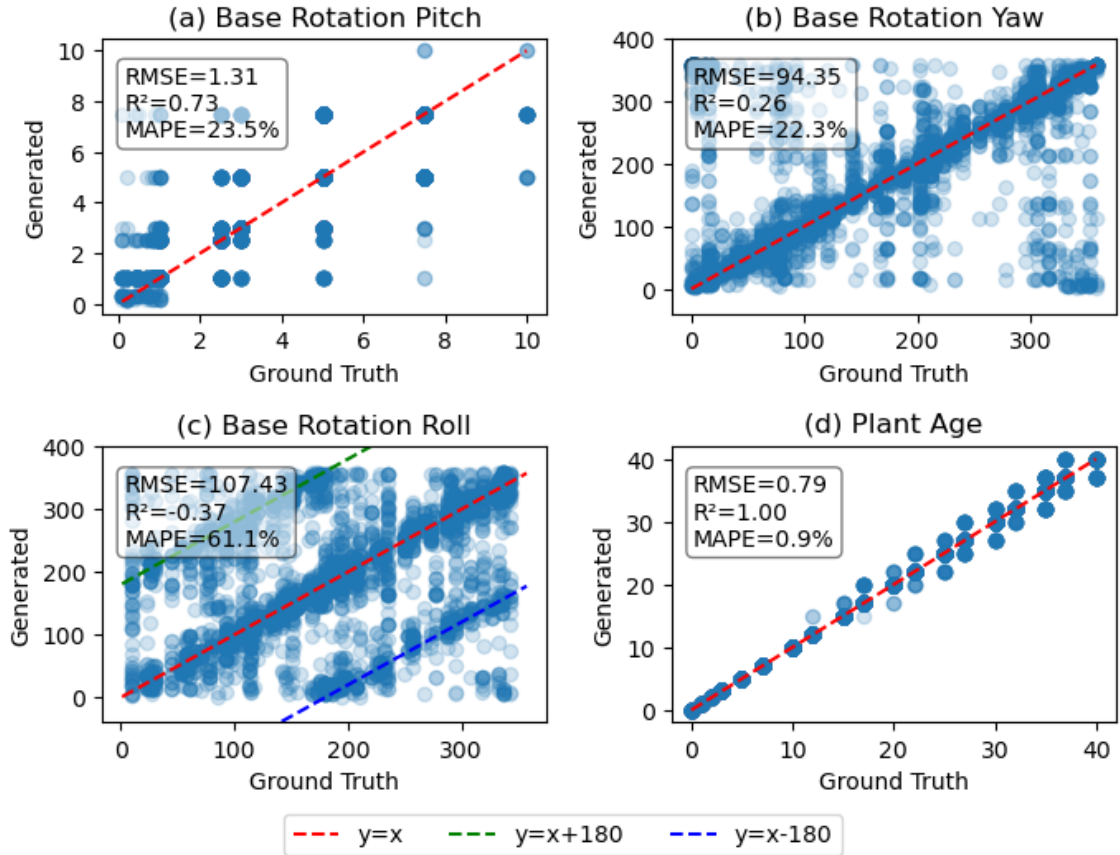

*Figure 7. Comparison of base rotation angles (pitch, yaw, and roll) of the first unifoliate shoot and plant age from the generated token sequence. The tokens for base rotation and plant age can be matched with the corresponding ground-truth plant architecture token sequence and converted to decimal values for comparison.*

As mentioned previously, the number of plant organs can be an important trait for plant phenotyping. Therefore, the numbers of plant organs from the generated plant architecture model and the ground truth were compared by counting plant organ tokens. The number of shoots refers to the number of shoot tokens in the plant architecture token sequence, which is a signal for creating a new branch. The number of phytomers was counted by grouping tokens that appear in the order of internode, petiole, and leaf into a single phytomer.

Comparing the total number of plant organs revealed a high correlation between the generated and the ground truth plant architecture, similar to the correlation observed with plant age. Fig. 8(a) shows the scatter plot of the number of shoots, with an $R^2$ value of 0.83 and an RMSE of

1.16. However, when compared to the 1:1 line, it was observed that the number of shoots predicted by the model was 20.7% less than the ground truth. Even if the generated plant architectures have fewer shoot tokens, as shown in Fig. 8(b), the model remains accurate in terms of the total number of phytomers in the plant architecture model.

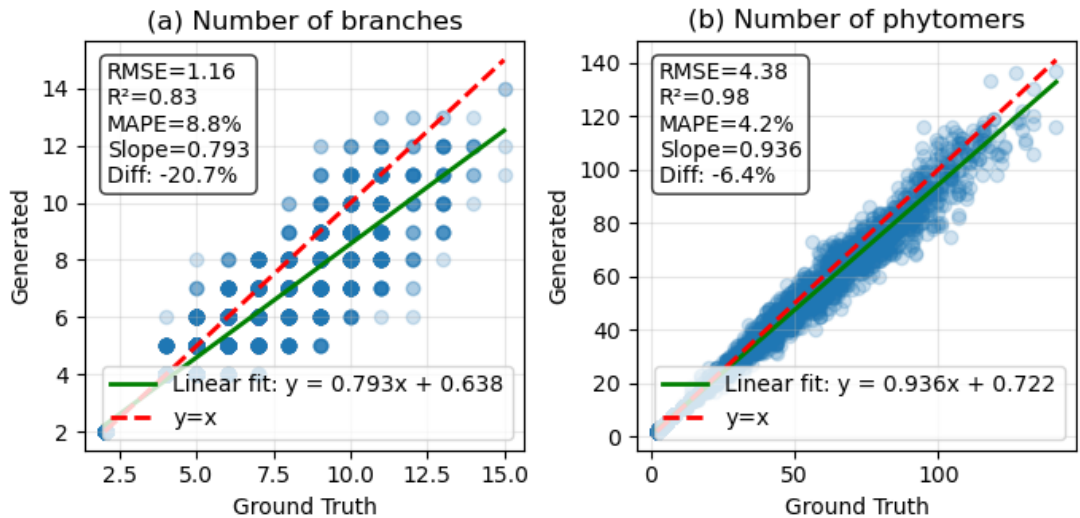

Figure 8. *Comparison of the (a) number of branches and (b) phytomers from the generated token sequence. Ground truth refers to the true counts in the dataset, and generated refers to the result obtained from the generated token sequence. Green lines represent the fitted linear regression between the ground truth and the generated data, while red dotted lines represent the y=x fit. Based on the slopes of the linear fit, the slope differences are displayed as Diff, showing -20.7% for (a) number of branched and -6.4% for (b) number of phytomers.*

The distributions of plant organ parameter tokens from generated plant architecture tokens and ground truth are illustrated in Fig. 9. As shown in Fig. 7, the model could estimate primary branch rotation and plant age parameters; these estimated parameters showed good agreement with the ground truth distributions, with a normalized Wasserstein distance of less than 0.05. Parameters that determine the size of each plant organ, such as internode length & radius, petiole length & radius, and leaf scale distributions, were also found to follow the ground truth distribution closely. However, some of the parameters showed normalized Wasserstein distance higher than 0.05, such as internode phyllotactic angle (0.059) and leaf pitch (0.065).

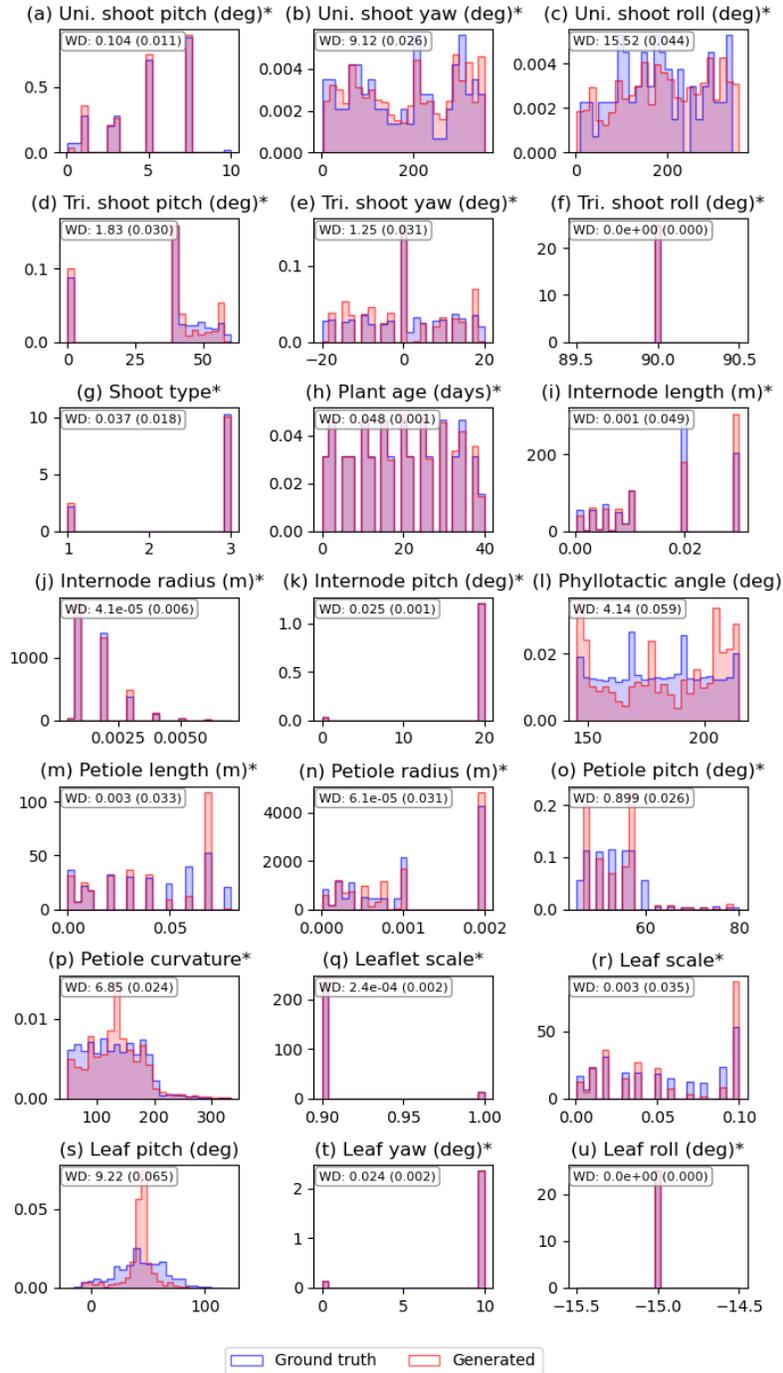

*Figure 9. Parameter distribution comparison from the test dataset (n=4,000) between the generated and ground truth. Uni. shoot refers to a branch where unifoliate leaves are connected, and Tri. shoot refers to a branch where trifoliate leaves are connected. The asterisk (\*) sign is labeled if the normalized Wasserstein distance (WD) is less than 5% (0.05).*

The leaf angle distribution of the plant architecture was calculated through geometric calculations based on leaf normals generated in the plant architecture simulation. A comparison was then made between the ground truth data and the model-generated leaf angle distributions. Fig. 10 shows the results of calculating the leaf Inclination angle using the plant architecture token sequence predicted from the image. Overall, the agreement ranged from a minimum of 0.018 to a maximum of 0.078 based on the normalized Wasserstein Distance. In the case of Fig. 10 (a) for day 5, the generated plant architecture predicted leaf angles in the 18° to 36° range and above 45° to 63° with a higher frequency than the ground truth, while predicting a lower frequency in the 36° to 45° and 63° to 90° ranges. For day 10 in Fig. 10(b), it predicted a higher frequency in the 18° to 36° range and similar or lower frequencies in the other ranges. In the cases of Fig. 10 (c) - (e), although some ranges were predicted with varying accuracy, the overall trend shows a normalized Wasserstein distance within 0.1, or a 10% difference.

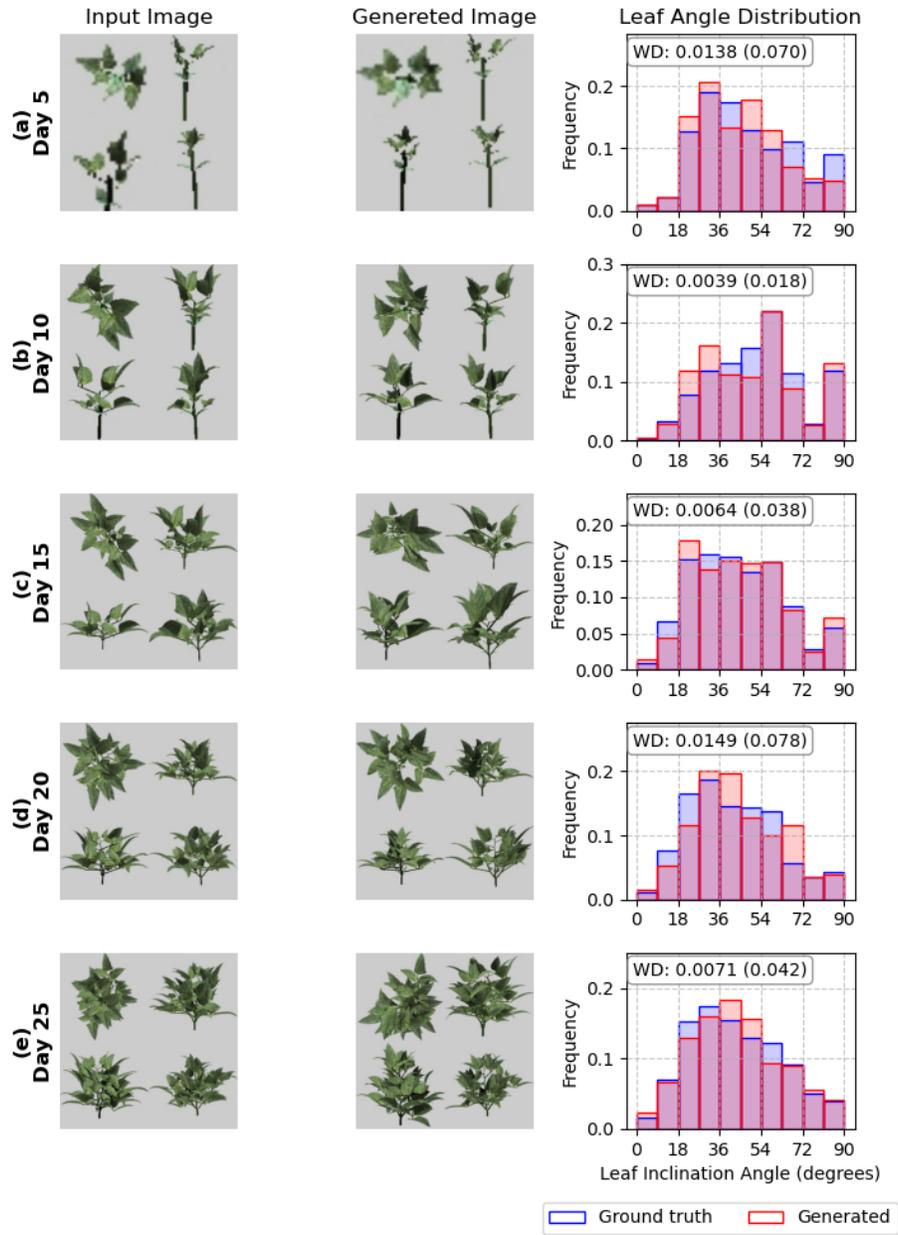

*Figure 10. Selected sample images from a test plant across the plant age dates (a-e). Left column: ground truth images from the dataset. Middle column: Re-rendered image from generated plant architecture. Right column: Comparison of leaf angle distribution between ground truth and generated, along with Wasserstein distance.*

# Plant height, leaf count, and leaf area estimation and comparison with naive baseline

This section calculates plant height, leaf count, and total leaf area using Helios plant simulation for 4,000 pairs of ground-truth and generated plant architecture token sequences, and compares them against a regression-based baseline method. Fig. 11 illustrates the results of a baseline model utilizing vision transformer features with a fully connected layer to estimate plant height, leaf count, and leaf area. As shown in Fig. 11, the ViT-B + FC model, which directly predicts plant traits using features from ViT-B, yields mean absolute percentage error (MAPE) values of 3.5%, 26.0%, and 11.8% for plant height, leaf count, and leaf area, respectively. Our model, which is shown as ViT-B + LM decoder, showed a higher MAPE value of 4.9% for estimating the plant height compared to the baseline method. However, leaf count and leaf area showed lower MAPE values than the baseline method, showing 4.1% for leaf count and 3.2% for leaf area.

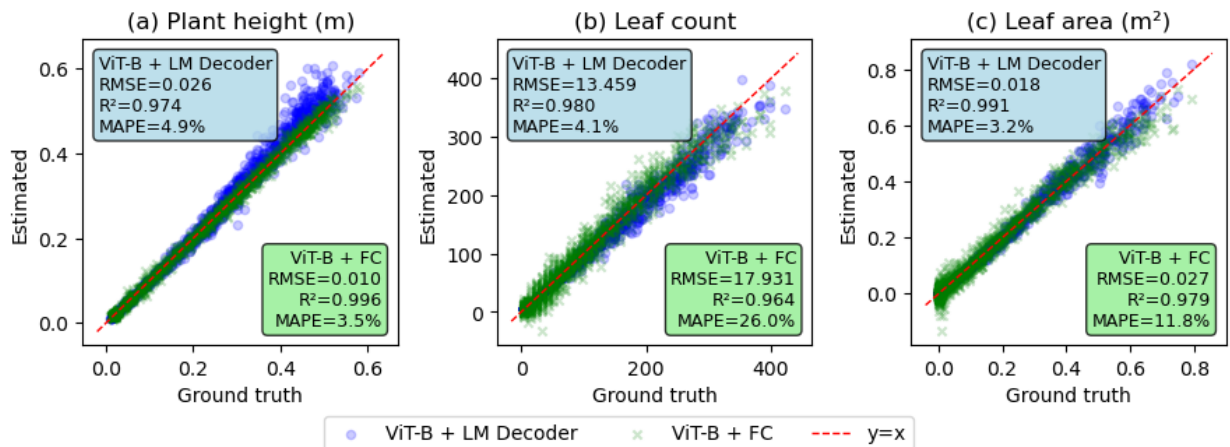

Figure 11. Comparison of plant scale bulk parameters (a) plant height, (b) leaf count, and (c) leaf area, derived from plant architecture generation and direct prediction using image features.

# Discussion

The main hypothesis of this paper was whether the details of plant architecture and topology can be learned from images using a large language model. We tested this hypothesis by using synthetic image data alongside detailed plant architecture information and training a vision-based large language model algorithm. Research predicting plant structure through images has been ongoing; however, this is the first study to utilize a vision language model to predict tokens that define plant structure. This study demonstrated that a vision language model can generate plant architecture, from organ-level to plant-level, from an image by estimating the tokens that build plant architecture. As shown in Fig. 6, even if the generated plant architecture overall appeared to reflect the general shape and complexity of the ground truth, it could not perfectly replicate the input image itself. This is because the model is only trained to predict the $n^{th}$ token from a sequence of tokens up to n-1 during the model training process. In other words, the model cannot access the rendered images generated from plant architecture tokens, and therefore, it has not been optimized to maximize the similarity of these images. Also, the generated plant architecture tends to have more branches than the ground truth.

Even though there is still room for improvement in the generated plant architecture, the results overall appeared to successfully estimate the plant traits from the input image. Fig. 7, 8, 9 showed that the plant architectural parameters can be estimated from the generated plant architecture. Still, there were some limitations in estimating the plant organ parameters. First of all, converting floating-point value parameters to integer token IDs introduced quantization error, transforming the parameter distributions into discrete distributions and changing the shape of the distribution. For example, the unifoliate shoot's pitch angle and internode phyllotactic angle were sampled from uniform distributions (Table 1), but the tokenized distributions (Fig. 9(a, i)) were not uniformly distributed. Secondly, a phenomenon called mean collapse occurred, where the data tends to cluster around the most frequently occurring values in the ground truth data, resulting in predictions collapsing to a few representative tokens. This is noticeable in the leaf parameters, such as in the leaf pitch values from Fig. 9 (s), where up to 400 leaves can appear in a single plant architecture model. Still, the overall distributions of estimated plant organ parameters followed the parameter distributions of the ground truth, showing a maximum value of normalized WD as 0.065.

By analyzing the generated plant architecture at the plant scale, indirect traits such as leaf angle distribution can be derived, as shown in Fig. 10, that can only be calculated from the 3D simulation and were not used for training explicitly. Although the leaf pitch angle distributions showed limited representation, the final calculated leaf angle distribution can be predicted with relatively high accuracy, showing normalized WD values of 0.018 to 0.078 (Fig. 10). This indicates that while the model cannot perfectly predict the angle of every individual leaf, it can accurately capture the overall trend of leaf angle distribution in the image without being explicitly trained to do so.

In addition to plant organ parameters, plant-scale bulk parameters, including plant height, leaf count, and leaf area, were compared between the predicted and the ground truth. Even if the baseline model only utilized a pretrained feature extractor with one layer of a fully connected layer, it could successfully predict those bulk parameters with high $R^2$ values, by explicitly learning those traits from image features. However, the baseline approach is limited to predicting individual traits and cannot capture the overall complex architecture of the plant, which is needed for FSPM simulation inputs. Compared to this, the plant architecture-based model implicitly learned plant height, leaf count, and leaf area through the plant architecture token sequence, and it was able to predict these variables more accurately than models trained directly on them. Notably, leaf count (Fig. 11 (b)) and leaf area (Fig. 11 (c)) could be predicted with higher $R^2$ and lower RMSE than direct estimation. The better accuracy on leaf count and leaf area means that the model, trained using plant architecture information, can more accurately predict traits that are difficult to estimate due to occlusion. In other words, this indicates that our model has learned the complex internal architecture of the plant—such as leaf count and leaf area—that cannot be easily predicted from 2D images without understanding plant architecture, enabling it to estimate traits closely related to the plant's structure.

# Conclusions

This research hypothesized that organ-level plant architecture can be generated from implicit information in plant images through a vision-language model. The results showed that a vision encoder-decoder model trained on a synthetic cowpea dataset could generate organ-level plant architectural reconstructions from images. Also, the model was capable of estimating whole-plant traits such as plant height, leaf count, and leaf area, approaching directly trained models' performance even if the language model never saw those values during training. The main takeaway of this paper is that plant architecture can be processed and tokenized for training with a large language model. The benefit of generating a synthetic dataset was that it decreased the ground truth dataset generation time relative to manual measurement in the field. Moreover, the synthetic dataset's accurate and comprehensive plant architecture information allows us to verify the algorithm's validity with "exact" ground truth data. Future research will extend this study to real image domains. We will improve this result by collecting real plant architecture data from the field to compare with our model's predictions, and improve performance using few-shot learning with the collected data. Also, we will test the algorithm with various backgrounds, lighting conditions, and occlusions to overcome the complexity of the actual environment.

The methodology used in this work has limitations that will be the target for improvement in future iterations. The current model training is an open-loop process, which means the rendered image from the generated plant architecture cannot provide feedback to the model weights. Therefore, future research should provide feedback on the generated image through reinforcement learning by adding the Helios simulation program in the training loop. We plan to incorporate reinforcement learning based on rendering image similarity into this pre-trained plant architecture token generation model, aiming to fine-tune the model so that the final rendered image is similar to the input image. Additionally, at present, only the plant organs of the shoot, internode, petiole, and leaf are considered, assuming the plants are in the vegetative growth stage; however, reproductive organs will be included in future studies.

In conclusion, this paper marks an initial step showing that a vision-based large language model can shift the paradigm for plant architecture modeling in plant science. By demonstrating this possibility with a synthetic dataset, this paper provides a foundation for the next step of applying our model to a larger dataset or a real environment.

# Acknowledgements

This work was supported, in whole or in part, by the Bill & Melinda Gates Foundation INV-0028630. Under the grant conditions of the Foundation, a Creative Commons Attribution 4.0 Generic License has already been assigned to the Author Accepted Manuscript version that might arise from this submission.